\documentclass[lettersize,journal]{IEEEtran}
\usepackage{amsmath,amsfonts}
\usepackage{algorithmic}
\usepackage{subfigure}
\usepackage{stfloats}
\usepackage{url}
\usepackage{verbatim}
\usepackage{threeparttable}
\usepackage{algorithm}
\usepackage{multirow} 
\usepackage{amsmath}
\usepackage{tcolorbox}
\usepackage{xcolor}
\usepackage[table]{xcolor}

\usepackage{booktabs}
\usepackage{graphicx}
\usepackage{cite}
\usepackage{tikz}
\usepackage{pgfplots}	
\usetikzlibrary{arrows.meta}
\usepgfplotslibrary{groupplots}
\usepackage{bm}
\usepackage{mathrsfs}
\usepackage{amsthm}
\usepackage{amsfonts}
\usepackage{amssymb} 
\usepackage{textcomp}
\usepackage{array}
\usepackage{utfsym}

\makeatletter
\renewcommand{\maketag@@@}[1]{\hbox{\m@th\normalsize\normalfont#1}}%

\usepackage{balance}
\pgfplotsset{compat=1.18}
\begin{document}
\title{Plug-and-Adapt: Multimodal Coreference Resolution at First Sight with a Pretrained Alignment Model}
\author{Jinghan~Wu,
 Jing~Li,
 Ivor W. Tsang~\IEEEmembership{Fellow,~IEEE},
 Xuetao~Zhang
\thanks{This research is supported by the National Research Foundation, Singapore under its National Large Language Models Funding Initiative (AISG Award No: AISG-NMLP-2024-003). Any opinions, findings and conclusions or recommendations expressed in this material are those of the author(s) and do not reflect the views of National Research Foundation, Singapore. This research is also supported in part by the National Natural Science Foundation of China (No: 62403372), the Science and Technology Project of China Huaneng Group Co., Ltd. (No: HNKJ23-HF67), the National Postdoctoral Innovative Talents Support Program (No: BX20230283), the ``Sanqin Bochuang" Talent Support Plan of Shaanxi Province (No: 2024SQBC005), and the China Scholarship Council (No: 202406280309). (\textit{Corresponding Author: Jing Li})}
\thanks{Jinghan Wu and Xuetao Zhang are with the State Key Laboratory of Human-Machine Hybrid Augmented Intelligence, the Institute of Artificial Intelligence and Robotics, Xi'an Jiaotong University, Xi'an 710049, China (e-mail:
wujing4022@stu.xjtu.edu.cn; xuetaozh@xjtu.edu.cn)}
\thanks{Jing Li and Ivor W. Tsang are with the Centre for Frontier AI Research and the Institute of High-Performance Computing, Agency for Science, Technology and Research (A*STAR), Singapore 138634
(e-mail: kyle.jingli@gmail.com; ivor.tsang@gmail.com)}}

\IEEEpubid{%
\begin{minipage}{\textwidth}
\scriptsize\centering
\textcopyright~2026 IEEE. All rights reserved, including rights for text and data mining and training of artificial intelligence and similar technologies. Personal use is permitted,\\
but republication/redistribution requires IEEE permission. See
https://www.ieee.org/publications/rights/index.html for more information.
\end{minipage}%
}

\markboth{IEEE TRANSACTIONS ON MULTIMEDIA}%
{How to Use the IEEEtran \LaTeX \ Templates}

\maketitle

\begin{abstract}
Visual information helps resolve ambiguity in coreference resolution, leading to notable performance gains. However, existing Multi-modal Coreference Resolution (MCR) methods require training with (partially) annotated data from the target dataset before they can be applied, preventing their direct usability and raising concerns about generalization. While Vision-Language Large Models (VLLMs) with billions of parameters offer promising zero-shot capabilities, they remain largely inaccessible. Their massive size limits deployability, and many are only accessible through paid APIs. In this paper, we propose a plug-and-adapt method that strategically adapts a carefully pre-trained \emph{alignment model} for immediate use in MCR tasks, designed to eliminate the need for training on scarce benchmark datasets or relying on resource-intensive VLLMs. Specifically, we first pre-train a fine-grained alignment model between textual and visual contextual information using vision-language alignment datasets. We then repurpose the alignment model to MCR through similarity aggregation by fusing visual and categorical cues with evidence theory, thereby enhancing effectiveness. Experiments on the Coreference Image Narratives (CIN) benchmark dataset demonstrate the effectiveness of our method, achieving a 5.31\% and 2.12\% improvement in CoNLL F1 over SOTA dedicated methods and popular VLLMs, respectively. We further evaluate our method on a masked CIN dataset for robustness testing and on a specially constructed VCR-MCR dataset for generalization assessment, with results confirming both capabilities.

\end{abstract}


\section{Introduction}\label{sec:intro}
Coreference resolution, the task of clustering mentions of the same entity, presents a fundamental NLP challenge due to linguistic ambiguity and complex contextual 
dependencies~\cite{liu2023brief}. While recent language model advancements have significantly improved coreference resolution ~\cite{joshi2020spanbert,kirstain2021coreference}, real-world scenarios like image captioning and storytelling~\cite{zhu2022image,singh2024pixels,shi2025multi,li2019video} often involve multiple modalities, necessitating multimodal coreference resolution (MCR) studies. For example, in the illustrated example shown in Fig.~\ref{fig:framework}, a text-only model alone may struggle to tell whether \emph{the woman} who is wearing a watch refers to \emph{a person} or \emph{another person} in the text, while it can be disambiguated after visually inspecting the paired image. 

\begin{figure}[!t]
 \centering
 \includegraphics[width=1
 \linewidth]{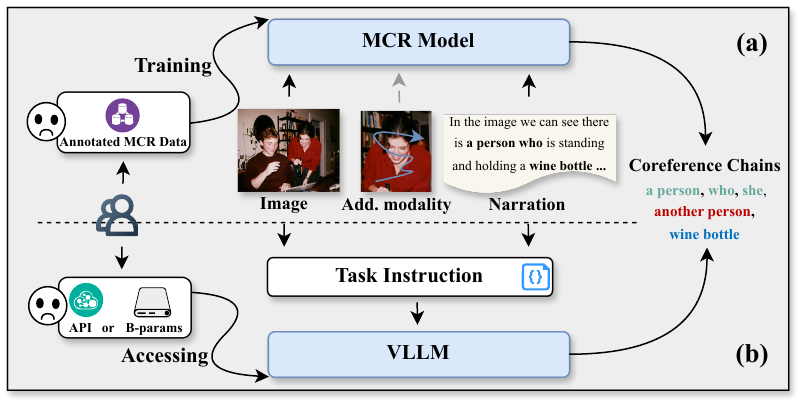}
 \caption{Current MCR solutions: (a)~dedicated MCR models, which require annotated data for training and subsequent use, and (b)~VLLMs, which rely on extensive computational resources for deployment or costly pay-to-use APIs.}
 \label{fig:Problem}
\end{figure}

Prior related efforts~\cite{kong2014you,cui2021s,xie2025phrase,guo2022gravl} have focused mainly on aligning textual phrases with visual regions, a fundamental task in vision-language modeling known as grounding~\cite{subramanian2022reclip,wang2023dual}, while largely overlooking coreference resolution. In addition, these studies have been restricted to specific domains, including indoor home environments~\cite{kong2014you}, Wikimedia~\cite{cui2021s}, and marketing conversations~\cite{kottur2021simmc}, limiting their applicability to restricted object categories and expression types. A recent benchmark dataset, Coreference Image Narratives (CIN)~\cite{goel2023you}, addresses these limitations by introducing open-world long-form narratives with annotated coreference chains and corresponding bounding boxes, serving as the primary benchmark for existing MCR models~\cite{goel2023semi,zheng2024self}. These methods are typically trained and evaluated on the CIN benchmark and rely heavily on its scarce and labor-intensive annotated data, specifically coreference chains, or on auxiliary information such as mouse tracking~\cite{pont2020connecting,goel2023you}~(Fig.~\ref{fig:Problem}(a)). \IEEEpubidadjcol
This prevents their direct usability and raises concerns about generalization. Most recently, Vision-Large Language Models (VLLMs) have  emerged as unified solutions for various tasks~\cite{lu2025revisiting,gao2024knowledge}, offering a promising alternative to dedicated methods through their zero-shot capabilities. However, these models face deployment challenges due to their massive size, with many being accessible exclusively through paid APIs (Fig.~\ref{fig:Problem}(b)). This raises a compelling question: \textbf{Is it possible to resolve coreference in image narrations without dedicated training on annotated coreference chains or resorting to VLLMs?}

 Although prior MCR studies~\cite{goel2023you,goel2023semi,zheng2024self}, along with our own experiments, suggest that directly applying VLMs~\footnote{We refer to vision-language models with fewer than one billion parameters as ``VLMs" and those with billions as ``VLLMs".} yields unsatisfactory results, their potential to align vision and language modalities is evident. In this paper, we rethink the MCR task, proposing a plug-and-adapt method that resolves coreference in image narrations by systematically adapting a pre-trained vision-language alignment model. Specifically, we first reorient the alignment model toward the referential complexity inherent in MCR through the relation formulation. We pre-train a relation-based grounding model built upon CLIP~\cite{radford2021learning} to capture the complex alignments between textual and visual entity interactions. These higher-order alignments prove more effective at resolving coreference ambiguities and linking previously unseen mentions that are prevalent in MCR but scarce in conventional grounding tasks, such as pronouns. Then, we bridge the gap between grounding and MCR demands by constructing informative mention representation. Since MCR demands modeling inter-mention similarity beyond simple alignment, the rigid one-hot grounding representations, biased toward top-aligned pairs, are ill-suited under alignment uncertainty. To address this, we design an aggregation method that leverages all relevant matching scores from the alignment model, yielding more informative mention representations. Finally, we consolidate the inference adaptation through multi-cue integration. Inspired by state-of-the-art dedicated methods~\cite{goel2023you,goel2023semi,zheng2024self}, we leverage multi-cue information from images, incorporating both visual and categorical features. To more effectively integrate these cues, we adopt evidence theory~\cite{sentz2002combination,josang2016subjective}, resulting in more reliable coreference resolution. Our contributions are as follows:

\begin{itemize}
 \item We are the first to tackle the MCR problem without relying on labor-intensive annotated data for dedicated training or on resource-intensive VLLMs, offering a benchmark-independent and accessible solution. Our work analyzes the key challenges of this task and lays the foundations for future research in coreference resolution.
\item We repurpose a pre-trained alignment model for the MCR task through relation formulation and aggregation-based mention representation. In addition, we design a novel multi-cue integration method via evidence theory, which consolidates the reliable inference adaptation.
 \item Experiments on the CIN dataset show our model surpasses SOTA dedicated methods and popular VLLMs, improving CoNLL F1 by 5.31\% and 2.12\%, respectively. Results on masked CIN and our newly established VCR-MCR dataset further validate its robustness and generalization.
\end{itemize}

\section{Related Work}
\subsection{Multi-modal Coreference Resolution}
Coreference resolution in image narrations is a challenging task involving unconstrained narrations with open-domain mention vocabularies and diverse referring expressions. This complexity underscores the need for dedicated studies in multi-modal coreference resolution (MCR). To advance this field, the Coreference Image Narratives (CIN) dataset~\cite{goel2023you} was developed and has since become the primary benchmark for MCR tasks. Building upon this benchmark, a weakly-supervised MCR model (WS-MCR)~\cite{goel2023you} was developed that leverages unlabeled image-narration pairs, linguistic heuristics, and auxiliary information to achieve promising performance. This was further advanced by a semi-supervised method (Semi-MCR)~\cite{goel2023semi}, which incorporates tailored loss functions for both labeled and unlabeled data to enhance MCR performance. More recently, data scarcity of MCR has been partially mitigated through multimodal data augmentation using diffusion models to synthesize training samples from text-only datasets, combined with a self-adaptive mechanism for selecting high-quality samples~\cite{zheng2024self}. Despite these advances, these methods heavily rely on training on the scarce benchmark dataset, which hinders their direct use and raises concerns about their generalization capability.

\subsection{Visual Language Models}
 Recent advances in VLMs have greatly improved multi-modal understanding and reasoning. Early efforts, such as VisualBERT~\cite{li2019visualbert}, used a single-stream architecture to jointly encode images and text. UNITER~\cite{chen2020uniter} built on this by introducing multiple pre-training objectives such as masked language modeling, masked region modeling, image-text matching, and word-region alignment. Zhang et al.~\cite{zhang2021vinvl} shifted focus to visual feature quality, using stronger object detectors to extract richer regional embeddings. Nevertheless, prior works~\cite{goel2023semi} have demonstrated that these early VLMs struggled to generalize to MCR tasks without fine-tuning. Subsequent developments such as CLIP~\cite{radford2021learning} achieved impressive zero-shot performance by learning image-text similarity, a capability later enhanced by Han et al.~\cite{han2024zero} through improved structural alignment. However, these models still fail to match the performance achieved by well-trained, task-specific MCR models. More recently, the advent of large vision-language models (VLLMs)~\cite{claude3,li2024llava,bai2025qwen2} has offered a unified framework for diverse vision-language tasks~\cite{wu2023can,umeton2024gpt,xiong20253ur}. Despite these remarkable advances, their massive model sizes hinder practical deployment, and many remain accessible only via paid APIs, further posing challenges for accessibility~\cite{wu2023can,umeton2024gpt}.

\begin{figure*}
 \centering
 \includegraphics[width=\linewidth]{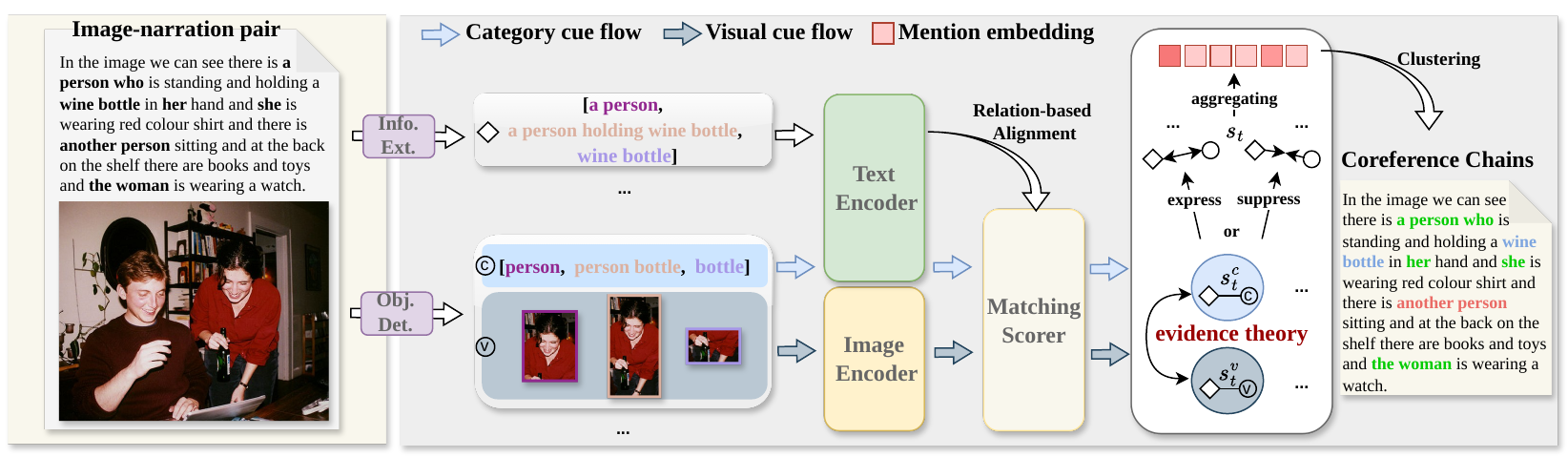}
 \caption{Overview of the adaptation stage of our method. Arrows with different colors in the diagram represent different cues.}
 \label{fig:framework}
\end{figure*}

\section{Methodology}
\subsection{Task Overview}
Suppose that we have an image-narration pair $(\mathcal{I}, \mathcal{N})$, where $\mathcal{I}$ is an image and $\mathcal{N}$ is the corresponding narration. The task of MCR is to cluster the unconstrained mentions in the narration that refer to the same entity depicted in the image. Unlike traditional text-only coreference resolution, MCR requires jointly reasoning over both visual and textual modalities. Formally, let $\mathcal{M} = \{m_{1}, m_{2}, \ldots, m_{|\mathcal{M}|}\}$ and $\mathcal{R} = \{r_{1}, r_{2}, \ldots, r_{|\mathcal{R}|}\}$ represent the set of mentions contained in $\mathcal{N}$ and the set of image regions in $\mathcal{I}$, respectively. We aim to obtain the similarity between any pair of mentions $m$ and $m^{\prime}$ in a given context of $\mathcal{N}$ with the support of $\mathcal{R}$, allowing us to resolve coreference.



\subsection{Pre-training with Relation-based Grounding}\label{PretrainVR}

Alignment data is relatively abundant~\cite{chao2015hico,pratt2020grounded,krishna2017visual} compared to MCR data, due to the complex annotation for coreference chains. We therefore begin by pre-training an alignment model to leverage this richer data. However, MCR is not merely a collection of isolated mention-region alignments, it also requires reasoning over referential dependencies and inter-mention relations. As a result, standard pairwise alignment alone is not fully sufficient for the downstream task.

To better bridge this gap, we reformulate alignment pre-training in a relation-aware manner. From a graph-based perspective, standard visual grounding can be viewed as bipartite matching~\cite{asratian1998bipartite,gao2022classification}, where textual expressions and visual regions form two types of nodes. For MCR, however, the key challenge lies not only in matching nodes across modalities, but also in capturing structured contextual dependencies within and across them. This is particularly important for unseen mentions such as pronouns, whose interpretation often depends on surrounding relational context.


An efficient way to introduce such structure is through relation triplets~\cite{gao2022classification,han2024zero}, which explicitly encode connections between nodes and support structured information propagation. This formulation preserves essential entity-relation information while remaining considerably accessible than alternatives such as full scene-graph modeling~\cite{yang2025llm}. We therefore adopt relational triplets in our framework.

Our pre-training process involves fine-tuning CLIP on a combination of three alignment datasets that contain rich relational knowledge~\cite{chao2015hico,pratt2020grounded,krishna2017visual} to enhance the relation-based alignment ability.
Specifically, given the textual description of the entities, referred to as mentions~\footnote{Here, ``mentions" refers specifically to noun phrases with one-to-one region mappings in grounding. In MCR tasks, the term extends to more reference types like pronouns and flexible mappings.}, and that of the relationship among entities, textual relation triplets can be formulated as:

\begin{equation}
t^{T}=(m_{i},p(m_{i},m_{j}),m_{j})
\label{eq:texttriplet}
\end{equation}
where $m_{i}$ denotes the subject mention, $m_{j}$ denotes the object mention ($m_{j}$ can be empty if there are no mentions related to $m_{i}$; for example, $t^{T}$ can be written as (\textit{a woman}, \textit{smiling}, \texttt{[blank]})), and $p(m_{i},m_{j})$ represents the relationship between the subject and object mentions. These datasets also provide visual regions for entities, allowing visual relation triplets to be formulated as:
\begin{equation}
 \begin{aligned}
 t^{I}=(r_{k},u(r_{k},r_{l}),r_{l})
 \end{aligned}
\label{eq:regiontriplet}
\end{equation}
where $r_{k}$ and $r_{l}$ denote the image regions cropped from the subject and object bounding boxes, respectively. The term $u(r_{k}, r_{l})$ denotes the union region covering both $r_{k}$ and $r_{l}$, which is used to represent the predicate between subject and object regions.

With relation triplets in narrations and images, we can now compute the triplet matching scores:
\begin{equation}
\begin{aligned}
s_{t}(t^{T},t^{I}) = \frac{1}{3} \big(&\cos(\mathbf{e}^{m_{i}},\mathbf{e}^{r_{k}}) + \cos(\mathbf{e}^{p_{i,j}},\mathbf{e}^{u_{k,l}}) \\
&+ \cos(\mathbf{e}^{m_{j}},\mathbf{e}^{r_{l}}) \big)
\end{aligned}
\label{eq:matchingvcue}
\end{equation}
where $\mathbf{e}^{m_{i}}$, $\mathbf{e}^{p_{i,j}}$, and $\mathbf{e}^{m_{j}}$ are the CLIP-encoded embeddings of the subject mention, predicate mention, and object mention, respectively, and $\mathbf{e}^{r_{k}}$, $\mathbf{e}^{u_{k,l}}$, and $\mathbf{e}^{r_{l}}$ are the CLIP-encoded embeddings of their corresponding subject, predicate, and object regions. We treat matching triplets as positives and others as negatives, applying standard contrastive loss to fine-tune CLIP for relation-aware grounding:
\begin{equation}
\begin{aligned}\sum_{(t_i^T,t_j^I)}\left[\log\left(\frac{s_t\left(t_i^T,t_j^I\right)}{\sum_ks_t\left(t_i^T,t_k^I\right)}\right)+\log\left(\frac{s_t(t_i^T,t_j^I)}{\sum_ks_t(t_k^T,t_j^I)}\right)\right]\end{aligned}\end{equation}

\subsection{Adaptation with Multi-cue Information} 
After obtaining the pre-trained alignment model, it would ideally be applied directly to the MCR task if the alignment model were an oracle. However, in real-world scenarios, oracle models are unattainable and gaps persist between grounding and MCR tasks. Hence, further adaptation is necessary.

\subsubsection{Bridge the Gap Between Grounding and MCR}\label{gap}
Information extraction (IE) is a fundamental task that involves extracting structured triplets in the form (subject, predicate, object) from text~\cite{grishman2015information,kamp2023open}. This process enables the generation of textual relation triplets for each narration, which can be represented as:
\begin{equation}
 \mathcal{T}^{T} = \{t^{T}=(m_{i},p(m_{i},m_{j}),m_{j})|1 \leq i,j \leq |\mathcal{M}|\}
\end{equation}

Note that $|\mathcal{T}^{T}|<|\mathcal{M}|^{2}$, as triplets are constructed only for mention pairs that exhibit a relationship. For image triplets, we first detect regions using an object detector~\cite{ren2015faster}. As inter-region relations are unknown, we construct all possible region pairs via Cartesian product. Following the pre-training process, we define the union of subject and object regions as their visual predicate, forming image relation triplets:
\begin{equation}
 \begin{aligned}
 \mathcal{T}^{I} = \{t^{I}=(r_{k},u(r_{k},r_{l}),r_{l})|1 \leq k,l \leq |\mathcal{R}|\} \\
 \end{aligned}
\label{eq:regiontripletset}
\end{equation}

These relation triplets help capture referential and contextual structures that cannot be effectively modeled by simple region-text alignment alone. After constructing the triplets, the similarity between a mention $m$ and a region $r$ can be computed by the similarities of their relevant top-aligned triplet pairs~\cite{han2024zero}. However, this bias toward self-assumed best alignment in grounding usually leads to inferior coreference accuracy, especially when grounding confidence is low. Therefore, in MCR, we do not stop at finding the best alignment but focus on learning informative representation for mentions.

Specifically, given a mention-region pair $(m, r)$, the set of triplet pairs $\mathcal{P}{(m,r)}$ associated with mention $m$ and region $r$ is formed as the union of triplet pairs in which $m$ and $r$ act as the subject, or $m$ and $r$ act as the object:
\begin{equation}
\begin{aligned}
\mathcal{P}{(m,r)} =\{(t^{T},t^{I})|~\text{Sub}(t^{T})=m,~\text{Sub}(t^{I})=r\}\cup \\\{(t^{T},t^{I})|~\text{Obj}(t^{T})=m,~\text{Obj}(t^{I})=r\}
\end{aligned}
\label{eq:matchingpair} 
\end{equation}
where $t^{T} \in \mathcal{T}^{T}$, $t^{I} \in \mathcal{T}^{I}$, $\text{Sub}(\cdot)$ and $\text{Obj}(\cdot)$ identify the subject and object of mention or region triplets, respectively. Then we can measure the similarity between the mention $m$ and the region $r$ by:
\begin{equation}
 s(m,r) = \frac{1}{|\mathcal{P}{(m,r)}|} \sum_{(t^{T},t^{I})\in\mathcal{P}{(m,r)}} s_{t}(t^{T},t^{I})
 \label{eq:matchingmentionregion}
\end{equation}
where $s_{t}(t^{T},t^{I})$ has the same meaning as in Eq.~(\ref{eq:matchingvcue}). The embedding of mention $m$ is computed based on its similarity scores with all regions, formulated as:
\begin{equation}
 f(m)=[s(m,r_{1}),s(m,r_{2}),\cdots, s(m,{r}_{|\mathcal{R}|})]
 \label{eq:mentionembedding}
\end{equation}
The similarity between mentions can then be calculated by:
\begin{equation}
 f_{cor}(m,m^{\prime}) = \frac{f(m) \cdot f(m^{\prime})}{{\Vert f(m) \Vert}{\Vert f(m^{\prime}) \Vert}}
 \label{eq:coreference}
\end{equation}

Following previous MCR works~\cite{goel2023you,goel2023semi}, if $f_{cor}(m,m^{\prime})$ exceeds a threshold, the two mentions $m$ and $m^{\prime}$ are considered coreferent and assigned to the same coreference chain.

According to Eqs.~\eqref{eq:matchingmentionregion} and \eqref{eq:mentionembedding}, the reliability of mention representation is determined by the fidelity of matching score $s_t(\cdot,\cdot)$. Essentially, for a textual triplet, its similarity with non-matching\footnote{We use ``non-matching" to refer to all unaligned region triplet pairs, even if the aligned region is incorrectly grounded.} visual triplets is not explicitly modeled during pre-training--contrastive learning treats them as uniformly negative pairs. This raises concerns about accumulating unreliable similarity scores in Eq.~\eqref{eq:matchingmentionregion}, especially when the number of detected regions is large.

\noindent \textbf{Remark 1.} \emph{For a mention $m$ that appears in $a$ triplets, $\mathcal{P}(m, r)$ has a cardinality of $a|\mathcal{R}|$, among which $a(|\mathcal{R}| - 1)$ are non-matching triplet pairs.}

Hence, it becomes crucial to develop a non-heuristic approach to reduce the potential distortions, which remains a challenge, as no ground-truth triplet pairs are available for supervision during adaptation. 
\subsubsection{Multi-cue Integration via Evidence Theory}\label{sec:multi-cue}
To consolidate the inference adaptation, our idea is to introduce another region-related cue to inspect whether the triplet matching in Eq.~\eqref{eq:matchingmentionregion} is reliable. Inspired by current state-of-the-art MCR methods that concatenated multiple cues of image regions for training~\cite{goel2023you,goel2023semi,zheng2024self}, we only utilize the region's category as an additional cue and construct the sets of category relation triplets in the same way as in Eq.~\eqref{eq:regiontripletset}. For ease of reference, their relation triplets are denoted as:

\begin{equation}
 \begin{aligned}
  \mathcal{T}^{{I}_{v}} = \{t^{{I}_{v}}=(r^v_{k},u(r^v_{k},r^v_{l}),r^v_{l})|1 \leq k,l \leq |\mathcal{R}|\} \\
 \mathcal{T}^{{I}_{c}} = \{t^{{I}_{c}}=(r^c_{k},u(r^c_{k},r^c_{l}),r^c_{l})|1 \leq k,l \leq |\mathcal{R}|\}
 \end{aligned}
\label{eq:regiontriplevc}
\end{equation}

Since both the image region $r^v$ and the textual category $r^c$ are detector outputs, there exists a one-to-one correspondence between $t^{{I}_{v}}$ and $t^{{I}_{c}}$. After obtaining triplets from different cues, we separately feed them into the model to compute their matching scores with respect to mention relation triplets, as shown in Fig.~\ref{fig:framework}. Then, our goal is to let two cues collaboratively compute the triplet matching, which can be done by applying the evidence theory-based fusion~\cite{sentz2002combination,josang2016subjective,xiao2020generalization}. Notably, our use of fusion here does not assume strict statistical independence between the visual and category cues. Instead, it only requires treating them as two distinct sources of evidence. In our work, the former is derived from region appearance, while the latter comes from detector-predicted semantic labels. Although both originate from the same image, they capture complementary information and different uncertainty characteristics, making this a practical approximation for uncertainty-aware fusion~\cite{han2022trusted}. Under this formulation, the fusion process is described below.

With the matching scores of the visual cue $s_{t}(t^{T},t^{I_v})$ and the category cue $s_{t}^{c}(t^{T},t^{I_c})$ ready to use, we first normalize the matching scores of two cues to the same scale by:
\begin{equation}\begin{aligned}
 \hat{s}_t^v(t^T,t^{I_v}) & = \frac{\left(s_t^v(t^T,t^{I_v})-\min\left(s_t^v(t^T,\cdot)\right)\right)}{\left(\max\left(s_t^v(t^T,\cdot)\right)-\min\left(s_t^v(t^T,\cdot)\right)\right)} \\
 \hat{s}_t^c(t^T,t^{I_c}) & = \frac{\left(s_t^c(t^T,t^{I_c})-\min\left(s_t^c(t^T,\cdot)\right)\right)}{\left(\max\left(s_t^c(t^T,\cdot)\right)-\min\left(s_t^c(t^T,\cdot)\right)\right)}
 \end{aligned}
 \label{minmax}
\end{equation}
where $s_t^v(t^T,\cdot)$ denotes the set of matching scores between $t^T$ and all triplets of the visual cue, and $s_t^c(t^T,\cdot)$ follows a similar definition. This pre-processing is indispensable, as the inter-modality distance in the embedding space of CLIP remains higher than the intra-modality distance, even after alignment tuning~\cite{schrodi2024two,yi2024bridge,shi2023towards}.

\begin{table*}[]
 \centering
  \caption{Coreference resolution (CR) results on the CIN dataset. Blocks 1-4: text-only models, dedicated MCR methods trained on the CIN dataset, VLMs without dedicated training on the CIN dataset, and VLLMs. Best and second-best results are bolded and underlined respectively. ``/" indicates unreported results in the original paper. ``Unk." indicates undisclosed parameter size.} 
 \setlength{\tabcolsep}{2.6pt}
 \begin{tabular}{lcccc|ccc|ccc|c}
 \toprule
\multirow{2}{*}{\textbf{Method}} & \multirow{2}{*}{\textbf{Params}}  & \multicolumn{3}{c}{\textbf{MUC}} & \multicolumn{3}{c}{\textbf{B$^{3}$}} & \multicolumn{3}{c}{\textbf{CEAF$_{\phi4}$}} & \textbf{CoNLL} \\ 
 \cline{3-12}\rule{0pt}{12pt}
 & & R (\%) & P (\%) & F1 (\%) & R (\%) & P (\%) & F1 (\%) & R (\%) & P (\%) & F1 (\%) & F1 (\%) \\
\midrule
Rule-Based &N/A& 5.60 & 10.13 & 6.40 & / & / & / & / & / & / & / \\
Neural Coref &10M& 0.11 & 0.17 & 0.13 & / & / & / & / & / & / & / \\
Longdoc & 370M & 7.79 & 8.43 & 7.24 & 62.27 & 76.10 & 67.69 & 48.77 & 84.95 & 61.02 & 45.31 \\ 
 Qwen2.5 & 8B & 39.63$_{\tiny \pm 1.63}$ & 39.49$_{\tiny \pm 1.53}$ & 36.67$_{\tiny \pm 0.73}$ & 82.72$_{\tiny \pm 1.12}$ & 85.76$_{\tiny \pm 0.95}$ & 83.66$_{\tiny \pm 0.98}$ & 70.56$_{\tiny \pm 1.35}$ & 87.82$_{\tiny \pm 0.79}$ & 77.46$_{\tiny \pm 0.68}$ & 65.93$_{\tiny \pm 0.35}$ \\ 
 \midrule
{MAF} &1.5M & 19.07 & 15.62 & 15.65 & / & / & / & / & / & / & / \\
{WS-MCR} &44M& 24.87 & 18.34 & 19.19 & / & / & / & / & / & / & / \\
{Semi-MCR} &209M & 31.11 & 35.25 & 31.86 & 70.63 & 87.85 & 78.06 & 63.99 & \underline{93.44} & 75.47 & 61.79 \\
{SA-MCR} & 209M & 39.83 & \textbf{44.70} & \textbf{41.43} & 72.42 & {90.27} & 80.32 & 66.51 & \textbf{96.63} & {78.38} & {66.71} \\ 
  \midrule

VisualBERT &123M&  18.17 & 6.08 & 8.06 & 69.01 & 36.08 & 41.03 & 21.25 & 57.10 & 28.67 & 25.92 \\
UNITER &303M&  16.92 & 7.15 & 8.83 & 68.34 & 44.29 & 50.22 & 28.12 & 72.78 & 38.91 & 32.65 \\
VinVL & 123M&   16.76 & 8.60 & 9.75 & 68.49 & 62.32 & 61.30 & 42.88 & 80.81 & 53.69 & 41.58 \\
CLIP &149M&   38.08 & 16.99 & 21.67 & 89.71 & 61.98 & 70.99 & 51.37 & 75.20 & 59.27 & 50.64 \\
VR-CLIP &149M&    33.28 & 22.72 & 25.20 & 88.28 & 77.19 & {81.60} & {69.02} & 82.04 & 74.29 & 60.36 \\
  \midrule
 Claude3 & Unk. & \textbf{49.67$_{\tiny \pm 1.60}$} & 39.66$_{\tiny \pm 1.29}$ & \underline{41.21}$_{\tiny \pm 0.54}$ & \underline{90.04}$_{\tiny \pm 0.61}$ & 85.50$_{\tiny \pm 1.05}$ & 86.58$_{\tiny \pm 0.46}$ & 76.86$_{\tiny \pm 1.16}$ & 88.17$_{\tiny \pm 0.91}$ & 80.91$_{\tiny \pm 0.19}$ & 69.57$_{\tiny \pm 0.13}$ \\
 LLaVA-OV & 8B &   0.20$_{\tiny \pm 0.14}$ & 17.59$_{\tiny \pm 1.33}$ & 11.17$_{\tiny \pm 0.83}$ & 75.83$_{\tiny \pm 1.34}$ & \textbf{94.88$_{\tiny \pm 1.16}$} & 83.75$_{\tiny \pm 1.21}$ & 77.21$_{\tiny \pm 1.06}$ & 82.38$_{\tiny \pm 0.89}$ & 78.80$_{\tiny \pm 0.78}$ & 57.91$_{\tiny \pm 0.94}$ \\
 QWen2.5-VL & 8B &  37.48$_{\tiny \pm 1.06}$ & 40.71$_{\tiny \pm 1.01}$ & 36.32$_{\tiny \pm 0.58}$ & 86.65$_{\tiny \pm 0.73}$ & \underline{92.41}$_{\tiny \pm 0.62}$ & \underline{89.00}$_{\tiny \pm 0.56}$ & \underline{83.11}$_{\tiny \pm 0.79}$ & 87.16$_{\tiny \pm 0.69}$ & \underline{84.38}$_{\tiny \pm 0.35}$ & \underline{69.90}$_{\tiny \pm 0.21}$ \\
  \midrule
\textbf{PA-MCR} &149M& \underline{41.79} & \underline{42.42} & 39.28 & \textbf{90.31} & 91.37 & \textbf{90.31} & \textbf{85.91} & 88.11 & \textbf{86.48} & \textbf{72.02} \\
\bottomrule
 \end{tabular}
  \label{tab:coreference}
 \end{table*}

After that, we can convert $\hat{s}_t^v(t^T,t^{I_v})$ and $\hat{s}_t^c(t^T,t^{I_c})$ into basic belief. For the pair of $(t^T,t^{I_v})$ with visual cue, we first define a binary frame $\Theta = \{A,\bar{A}\}$, where $A$ denotes the hypothesis that the matching state of $t^T$ and $t^{I_v}$ is true and $\bar{A}$ denotes the opposite. The power set of $\Theta$ is denoted as $2^{\Theta} = \{\emptyset,\{A\},\{\bar{A}\},U\}$. Here, $U=\Theta$ can be regarded as the proposition of uncertainty according to subjective logic~\cite{josang2016subjective}. In our work, we assume that cosine similarity provides evidence only for $A$ and $U$, i.e., $A$ and $U$ are treated as the focal elements (See Appendix~\ref{AAbS} for more discussion)~\cite{xiao2020generalization}. Based on this, we define the basic belief assignment for the visual cue that satisfies:
\begin{equation}
 \begin{aligned}
 & m^v_A(t^T,t^{I_v}) = \min\left(\hat{s}_t^v(t^T,t^{I_v}),\beta\right)  \\
 & m^v_U(t^T,t^{I_v}) = 1 - \min\left(\hat{s}_t^v(t^T,t^{I_v}),\beta\right)  \\
 \end{aligned}
 \label{eq:basicbeliefv}
\end{equation}
Here, $\beta$ serves as a safety margin, acting as a conservative cap when converting similarity into evidential belief, thereby preventing overconfident single-cue predictions while preserving uncertainty for subsequent fusion. This is motivated by the fact that similarity scores in contrastive learning reflect relative alignment rather than fully calibrated certainty~\cite{wu2024confidence,guo2017calibration}. 

Similarly, the basic belief assignments $m^c_A(t^T,t^{I_c})$ and $m^c_U(t^T,t^{I_c})$ are defined for the category cue in the same way. Based on the basic belief assignments of two cues, the combinational belief of $U$ in the Dempster-Shafer theory~\cite{sentz2002combination} is written as:
\begin{equation}
 m^{vc}_U(t^T,t^I) = \frac{m^v_U(t^T,t^{I_v})m^c_U(t^T,t^{I_c})}{1 - K}
 \label{eq:belief}
\end{equation}
where $K$ is the conflict. Since we only consider $A$ and $U$ here, the conflict can be considered zero. We then convert it into a normalized form:
\begin{equation}
 q(t^T,t^I) = \frac{e^{-m^{vc}_U(t^T,t^I)/\tau}}{\sum_{t_{*}^I \in \mathcal{T}^{I}} e^{-m^{vc}_U(t^T,t_{*}^I)/\tau}}
 \label{eq:uncertainty}
\end{equation}
which serves as a confidence weight for score aggregation. Here, $\tau$ acts as a temperature parameter in the Gibbs distribution. Smaller values of $\tau$ assign higher weights to low-uncertainty matches, whereas larger values lead to a softer and more evenly distributed weighting scheme. Therefore, $\tau$ provides explicit control over the balance between confident selection and robust aggregation.

Then, the updated form of mention-region similarity becomes:
\begin{equation}
 s^*(m, r) =\frac{1}{|\mathcal{P}{(m,r)}|}\sum_{(t^{T},t^{I})\in\mathcal{P}{(m,r)}} q(t^T,t^I)s_{t}(t^{T},t^{I})
 \label{eq:finalsimilarity}
\end{equation}

The above process serves as an ``express or suppress'' mechanism that enhances the matching scores of the reliable triplet pairs while suppressing those that are more uncertain. Building on this, the evidence-theoretic fusion module can integrate visual and category cues to recalibrate uncertain matches in the presence of noisy detections.

\noindent\textbf{Remark 2.} \emph{Note that the choice of $s_{t}(t^{T},t^{I})$ in Eq.~\eqref{eq:finalsimilarity} can be either $s_{t}^{v}(t^{T},t^{I_v})$ or $s_{t}^{c}(t^{T},t^{I_c})$, or even both. In our proposed plug-and-adapt method, based on the recent finding that information imbalance between modalities persists even with oracle encoders~\cite{schrodi2024two}, we would always take the visual cue as a support for the category cue by setting $s_{t}(t^{T},t^{I}) = s_{t}^{c}(t^{T},t^{I_c})$, as MCR remains a text-oriented task and the category cue is more semantically aligned with the textual modality than the raw visual region features.
} 

Subsequently, the mention embedding can be obtained by replacing $s(m,r)$ in Eq.~\eqref{eq:mentionembedding} with $s^*(m,r)$ and the final mention similarity can be computed by Eq.~\eqref{eq:coreference}. We refer to our complete method as PA-MCR hereafter for brevity.

\section{Experiments}
In this section, we conduct comprehensive experiments to address the following research questions: \textbf{RQ1}: Can PA-MCR provide a benchmark-independent and accessible solution for MCR? \textbf{RQ2}: To what extent is PA-MCR robust and generalizable? \textbf{RQ3}: How does multi-cue integration contribute to adaptation? and \textbf{RQ4}: How do upstream errors affect PA-MCR and when does it fail?

\subsection{Experimental Setup}
\subsubsection{Evaluation Metrics}
For the coreference resolution task, we evaluated the performance using standard coreference resolution metrics, including MUC~\cite{sundheim1995overview}, B$^{3}$~\cite{bagga1998algorithms}, CEAF$_{\phi4}$~\cite{luo2005coreference}, and their average F1 score, CoNLL F1~\cite{pradhan2012conll}. For the narrative grounding task, in line with previous work, we report the grounding accuracy for both noun mentions and pronouns. If the predicted bounding boxes have an Intersection over Union (IoU) greater than 0.5 with the ground-truth bounding boxes, we regard the prediction as correct.
\subsubsection{Baselines and Implementation}
We compare our method against four groups of baselines: text-only CR models, dedicated MCR models, VLMs, and VLLMs. Text-only methods perform coreference resolution directly on narrations without using images. This group includes the Rule-Based system~\cite{lee2011stanford} that relies on handcrafted mention rules, Neural Coref~\cite{lee2017end} trained on large general-purpose corpora, Longdoc~\cite{toshniwal2021generalization}, a transformer-based model built on Longformer-Large and trained on multiple text datasets, and Qwen2.5, a top-tier large language model that demonstrates competitive performance on a variety of NLP tasks~\cite{Qwen2024qwen2}. Multimodal methods leverage both narrations and images. For SOTA dedicated MCR methods, MAF~\cite{wang2020maf} is a weakly supervised grounding model fine-tuned on the CIN dataset, WS-MCR~\cite{goel2023you} enhances weak supervision through contrastive learning and linguistic priors, Semi-MCR~\cite{goel2023semi} combines labeled and unlabeled data in a semi-supervised framework, and SA-MCR~\cite{zheng2024self} achieves strong supervised performance with data augmentation. For VLMs, we conduct zero-shot evaluation by computing cosine similarity between multimodal mention embeddings, covering VisualBERT~\cite{li2019visualbert}, UNITER~\cite{chen2020uniter}, VinVL~\cite{zhang2021vinvl}, CLIP~\cite{radford2021learning}, and VR-CLIP~\cite{han2024zero}, which improves relation grounding. For VLLMs, we also adopt a zero-shot setting with a unified prompt adapted from SOTA LLM-based CR work~\cite{le2024language}, evaluating Claude3 (HaiKu)~\cite{claude3}, LLaVA-OV~\cite{li2024llava}, and QWen2.5-VL~\cite{bai2025qwen2}. Notably, given that different decoding behaviors may influence the performance of VLLMs, we evaluate VLLMs under different decoding temperatures, $\{0.0, 0.2, 0.6\}$, and report the mean and variance of the corresponding results. We also list the approximate parameter sizes of all models in Table~\ref{tab:coreference}.

For implementation, in our model, we use a 63M-parameter Transformer as the text encoder, consisting of 12 layers, a hidden width of 512, and 8 attention heads, together with a ViT-B/32 Transformer as the image encoder. This setup follows the architectures adopted in CLIP and VR-CLIP. During pre-training, we fine-tune the model on the union of three public datasets: HICO-det~\cite{chao2015hico}, SWiG~\cite{pratt2020grounded}, and Visual Genome~\cite{krishna2017visual}. For inference, we employ Faster-RCNN~\cite{ren2015faster} to extract visual regions and their corresponding category texts, consistent with previous dedicated MCR methods~\cite{goel2023you,goel2023semi,zheng2024self}. A shared object detector is used across all methods when necessary. We retain at most 100 detected boxes per image, discard boxes with detection confidence below 0.8, and remove boxes whose area is smaller than 0.5\% of the image area. We also use the same IE extractor~\cite{dong2023open} across all methods when necessary to ensure consistency. For textual triplets, we retain at most 28 detected relations per mention. Ground-truth mentions are provided for all multimodal baselines, following the settings used in previous dedicated MCR work~\cite{goel2023you,goel2023semi,zheng2024self}. For hyperparameter settings, unless otherwise specified, we set $\beta=0.9$ and $\tau=0.1$, and report all results under this setup.

\begin{table}[]
\centering
 \caption{Narrative grounding performance on the CIN dataset. * indicates models trained on the CIN dataset, which  includes pronoun annotations.}
  \setlength{\tabcolsep}{10pt}
\begin{tabular}{lcc|c}
\toprule
\textbf{Method} & \textbf{Noun Phrases} & \textbf{Pronouns} & \textbf{Overall} \\ 
\midrule
MAF$^{*}$ & 21.60 & 18.31 & 20.91 \\
WS-MCR$^{*}$ & 30.27 & 25.96 & 29.36 \\
Semi-MCR$^{*}$ & 32.58 & 28.45 & 31.71 \\
SA-MCR$^{*}$ & {36.84} & {32.14} & {35.72} \\
\midrule
CLIP & 17.13 & 0.06 & 16.14 \\
VR-CLIP & 32.26 & 22.83 & 31.44 \\
QWen2.5-VL & \underline{40.94} & \underline{46.42}& \underline{41.41}\\
\midrule
\textbf{PA-MCR} & \textbf{41.71} &  \textbf{49.68} &  \textbf{42.40} \\ 
\bottomrule
\end{tabular}
 \label{tab:grounding}
\end{table}
\begin{figure}[!t]
\centering
\includegraphics[width=\linewidth]{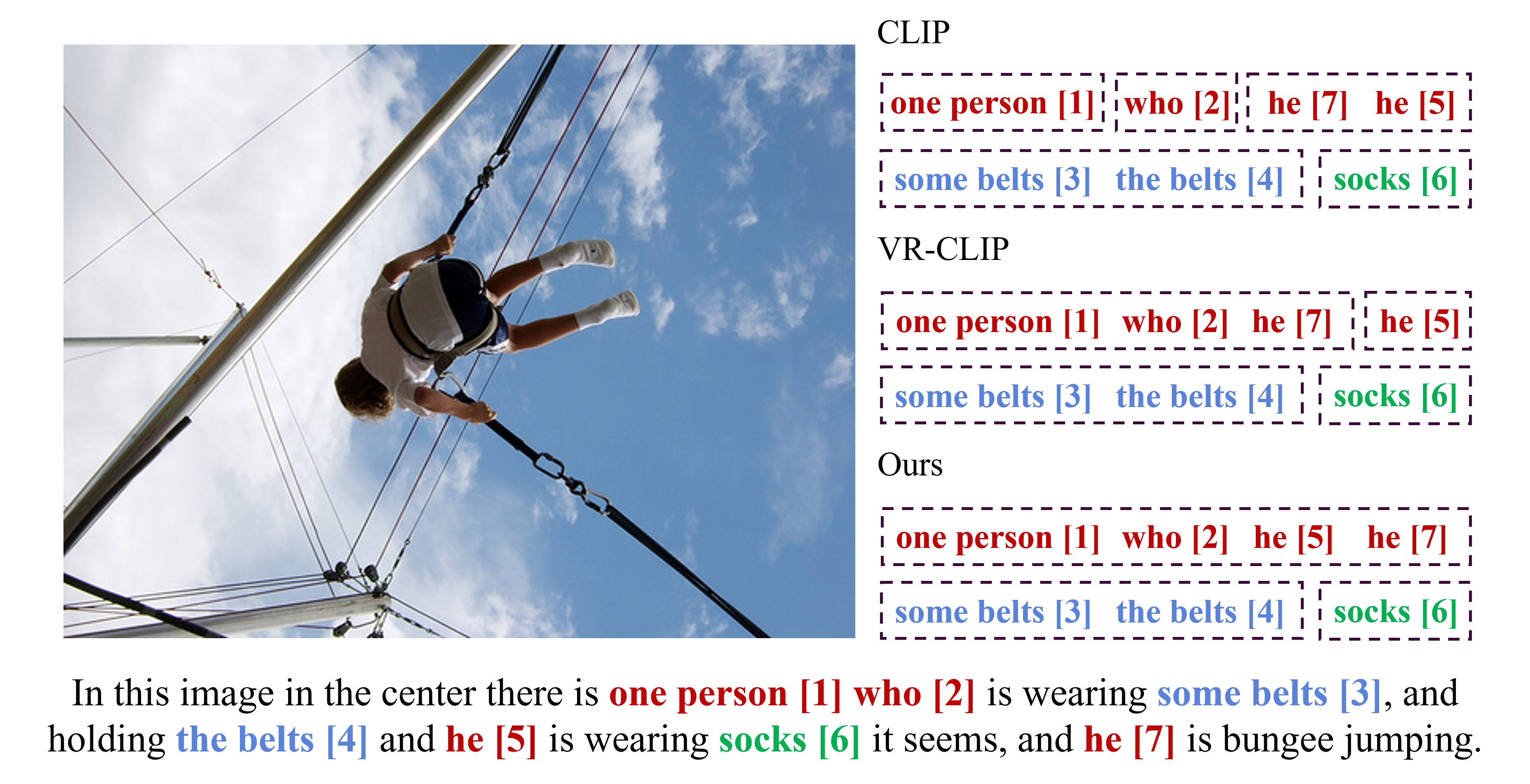}
\caption{A qualitative example showing coreference resolution result of CLIP, VR-CLIP, and PA-MCR.}
\label{fig:qualitativeanalysis}
\end{figure}
\subsection{Holistic MCR Comparison (RQ1)}\label{sec:CR_Comparison}
Table~\ref{tab:coreference} presents the coreference resolution results on the CIN dataset. Key observations include:
\subsubsection{Necessity of Visual Modality}
Under comparable model scales, text-only models (including Rule-Based, Neural Coref, Longdoc, and Qwen2.5) perform poorly in the absence of visual information, whereas multimodal models achieve significantly better results, highlighting the critical importance of the visual modality for MCR.

\subsubsection{Benchmark Independence}
As shown in the table, the performance of dedicated MCR methods is strongly influenced by the amount of annotated training data available. MAF and WS-MCR improve over text-only baselines, but their gains remain limited because they rely mainly on linguistic regularization and additional side information. Semi-MCR achieves substantial improvements by leveraging annotated CIN data, and SA-MCR further benefits from data augmentation. In contrast, our method is not trained on annotations from the target MCR benchmark (e.g., CIN coreference annotations), which is what we refer to as  benchmark independence. Even under this setting, our method still outperforms the strongest prior approach, SA-MCR, by 9.99\% in B$^{3}$ F1 and 5.31\% in CoNLL F1.


\subsubsection{Accessibility}
VLMs like VisualBERT, UNITER, VinVL, and CLIP exhibit notably poor performance on the MCR task. Notably, both VR-CLIP and our method were pre-trained with relation formulation on the same dataset. Under this shared setting and at similar parameter scales, our method achieves superior performance. Furthermore, compared with currently popular VLLMs with billions of parameters, our approach achieves competitive or even superior performance, improving the CoNLL F1 score while operating with significantly fewer parameters. For time complexity, the main additional cost of our design lies in the matching, aggregation, and mention-pair similarity computation, whose complexity is $O(|\mathcal{T}^T||\mathcal{R}|^2 + |\mathcal{M}|^2|\mathcal{R}|)$ after feature extraction. This cost is further mitigated in practice through parallel computation and region filtering, yielding an approximately $8\times$ speed-up in the matching overhead. These results indicate that the method remains practically accessible while delivering good performance.


\begin{figure}[!t]
    \centering
    \includegraphics[width=0.90\linewidth]{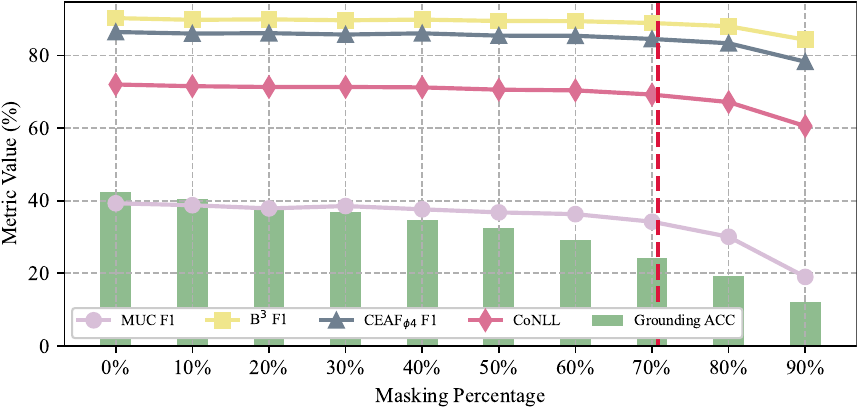}
    \caption{Grounding and CR results on the CIN dataset when randomly masking different percentages of image regions.}
    \label{fig:random_mask}
\end{figure}

\begin{table}[!t]
 \centering
 \caption{CR results (F1) on the VCR-MCR dataset.}
   \setlength{\tabcolsep}{10pt}
\begin{tabular}{lccc|c}
 \toprule
 \textbf{Method} & \textbf{MUC}& \textbf{B$^{3}$} & \textbf{CEAF$_{\phi4}$} & \textbf{CoNLL}\\
 \midrule
Semi-MCR   & 22.15 & 65.29 & 61.30 & 49.58 \\
SA-MCR     & 34.19 & 73.98 & 73.14 & 60.44 \\
VR-CLIP    & 30.69 & 74.26 & 63.13 & 56.03 \\
Claude3    & 31.78 & 73.23 & 61.59 & 55.53 \\
LLaVA-OV   & 9.42  & 59.23 & 49.76 & 39.47 \\
QWen2.5-VL & \underline{32.23} &  \underline{84.85} &  \underline{79.50} &  \underline{65.53} \\
 \midrule
 \textbf{PA-MCR} & \textbf{36.57} & \textbf{86.52} & \textbf{80.05} & \textbf{67.71} \\
 \bottomrule
 \end{tabular}
     \label{tab:newdata}
\end{table} 

\subsection{Robustness and Generalization Analysis (RQ2)}
\subsubsection{Study on Unseen Mentions}
To assess generalization to unseen mention types, we evaluated our method's narrative grounding performance (both nouns and pronouns) on the CIN dataset, comparing against dedicated MCR models and some VLMs/VLLMs. Table~\ref{tab:grounding} shows that while CLIP exhibits grounding capabilities for noun phrases, it fails with pronouns due to the absence of pronoun training data. VR-CLIP improves pronoun grounding through improved relationship modeling. Our method achieves superior performance on unseen mentions, outperforming dedicated grounding models and even VLLMs with strong zero-shot capabilities.
Furthermore, Fig.~\ref{fig:qualitativeanalysis} presents a qualitative example showing coreference resolution results of CLIP, VR-CLIP, and PA-MCR. A progressive improvement is evident across these models: CLIP poorly clusters pronouns, VR-CLIP performs better, and our method achieves optimal results. These results illustrate that our strategies of introducing relation triplets and evidence-theoretic fusion enable better generalization to unseen mention types.

\subsubsection{Reliance on Image Modality}\label{sec:Reliance}
We randomly masked 10\%-90\% of detected regions and evaluated model performance on masked images with original narrations. Fig.~\ref{fig:random_mask} shows results with bars indicating grounding performance and lines showing coreference resolution performance. Grounding accuracy drops rapidly with increased masking as target regions become more likely to be masked. However, coreference resolution remains stable below 70\% masking, demonstrating robustness with reduced visual information. Coreference performance drops sharply above 70\% masking, indicating mention embedding collapse. Interestingly, this threshold matches the average mention-to-region ratio (red dashed line), where visual information becomes insufficient.

\begin{figure}[!t]
    \centering
    \subfigure[CoNLL F1]{
    \includegraphics[width=0.46\linewidth]{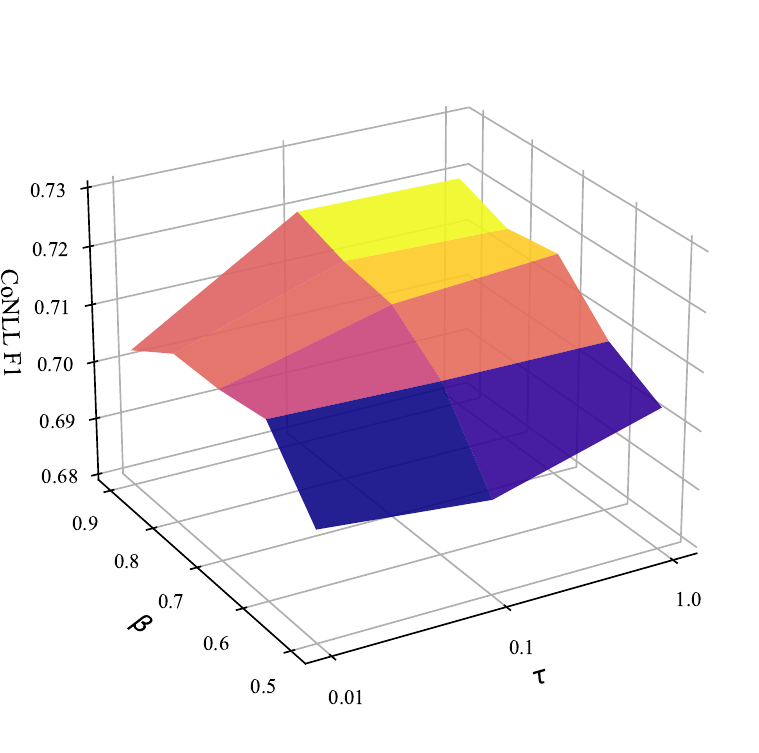}
        \label{fig:paramcr}
    }
    \hfill
    \subfigure[Grounding ACC]{
    \includegraphics[width=0.46\linewidth]{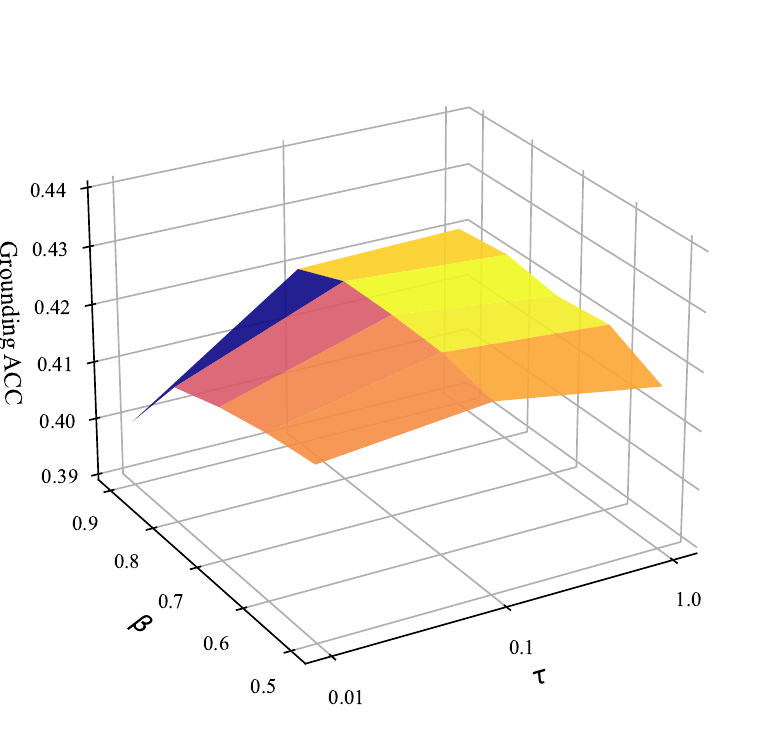}
        \label{fig:paramvr}
    }
    \caption{Parameter analysis for $\beta$ and $\tau$ on CoNLL F1 and Grounding ACC.}
       \label{fig:parameter_analysis}
\end{figure}

\subsubsection{Evaluation on Newly Constructed Data} 
\begin{figure}[!t]
    \centering
    \includegraphics[width=\linewidth]{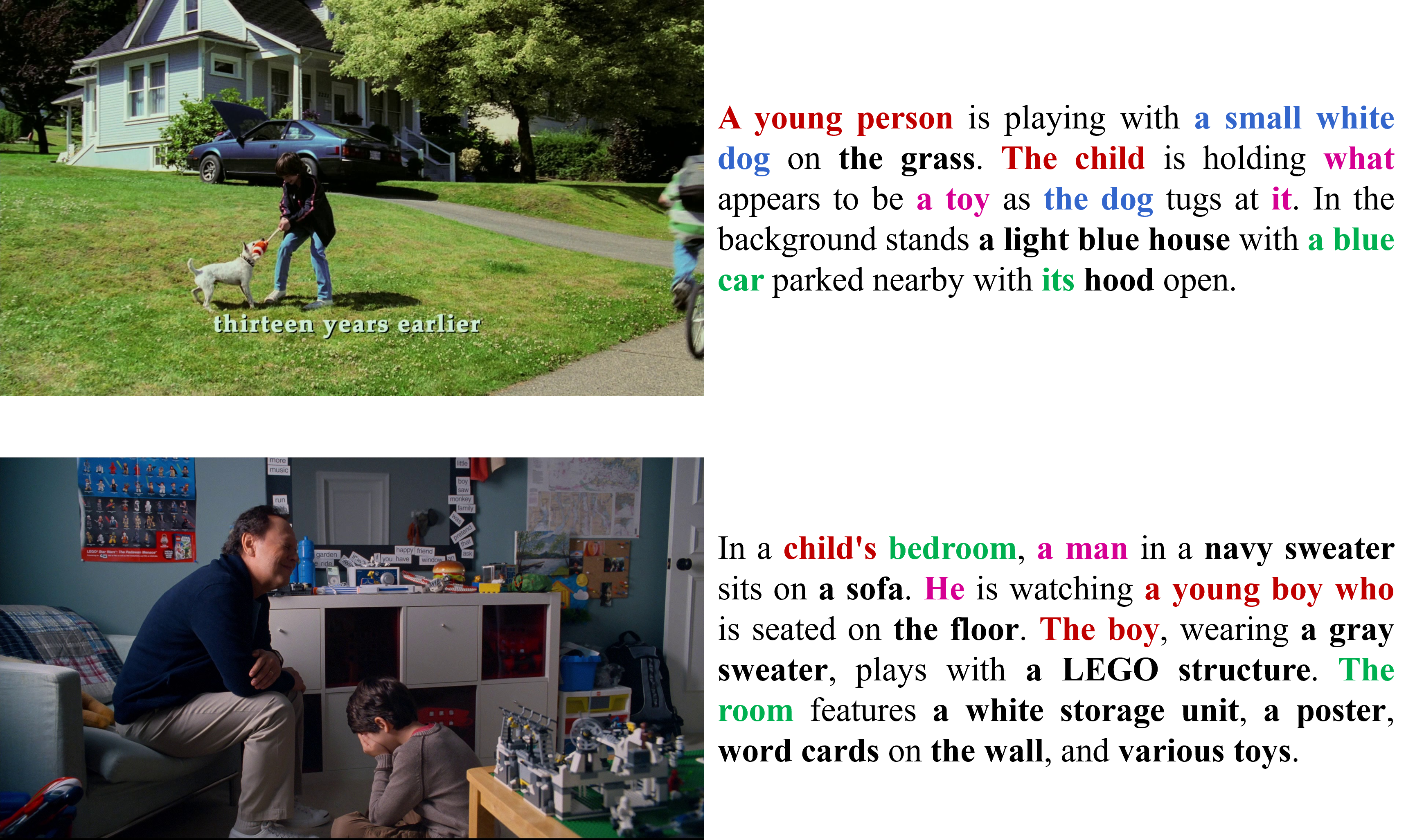}
    \caption{Examples of newly constructed data. The mentions in the same color form a coreference chain, and the singletons (coreference chains of length one) are highlighted in bold for simplicity.}
    \label{fig:newdata}
\end{figure}

In addition to the CIN dataset, we established a new self-annotated dataset, VCR-MCR, to address the scarcity of test beds for MCR. Specifically, we selected 900 images with diverse movie scenes from the Visual Commonsense Reasoning dataset~\cite{zellers2019recognition}. This dataset includes images drawn from movies with substantially different visual styles and themes, while the target entities span a broad range of object categories. For each image, we used an LLM to generate a corresponding narration containing diverse types of referring expressions, followed by manual verification. Subsequently, the coreference chains were manually annotated for the generated narration. During manual annotation and verification, we intentionally preserved ambiguous references whenever appropriate so that the data would better meet the objective of MCR. Some representative examples can be found in Fig.~\ref{fig:newdata}. To evaluate generalization, we tested our method on this dataset. The results in Table~\ref{tab:newdata} demonstrate that PA-MCR substantially outperforms VR-CLIP, validating cross-dataset effectiveness. Meanwhile, PA-MCR consistently outperforms the dedicated models, which is consistent with the conclusion drawn from the CIN results and further confirms its superior generalization ability.

\subsubsection{Parameter Sensitivity}
Our method involves two hyperparameters: the temperature $\tau$ in Eq.~(\ref{eq:uncertainty}) and the maximum confidence $\beta$ in Eq.~(\ref{minmax}). To assess their influence, we vary $\beta$ in $\{0.5, 0.6, 0.7, 0.8, 0.9\}$ and $\tau$ in $\{0.01, 0.1, 1\}$. The results for coreference resolution and grounding on the CIN dataset are shown in Fig.~\ref{fig:paramcr} and Fig.~\ref{fig:paramvr}. As illustrated in Fig.~\ref{fig:parameter_analysis}, CoNLL F1 generally increases with higher values of $\beta$, whereas grounding ACC is more sensitive to intermediate $\beta$ values. Both coreference resolution and grounding exhibit optimal performance around $\tau \approx 0.1$. Although the metrics show some variation across different parameter settings, the changes are relatively minor. This indicates that the proposed method is robust with respect to the choice of $\tau$ and $\beta$.

\begin{table}[!t]
 \centering
 \caption{Ablation study on PA-MCR with different cues.}
 \setlength{\tabcolsep}{10pt}
 \begin{tabular}{lcc|c}
 \toprule
\textbf{Metrics (F1)} & \textbf{Visual-cue} & \textbf{Category-cue} & \textbf{Multi-cue} \\
  \midrule
 MUC & 35.91 & \underline{38.24} & \textbf{39.28} \\
 B$^{3}$ & \underline{89.72} & 89.69 & \textbf{90.31} \\
 CEAF$_{\phi4}$ & 85.78 & \underline{86.09} & \textbf{86.48} \\
 CoNLL & 70.47 & \underline{71.34} & \textbf{72.02} \\ 
 \bottomrule
\end{tabular}
 \label{tab:ablation}
\end{table}
\begin{figure}[!t]
\centering
\includegraphics[width=\linewidth]{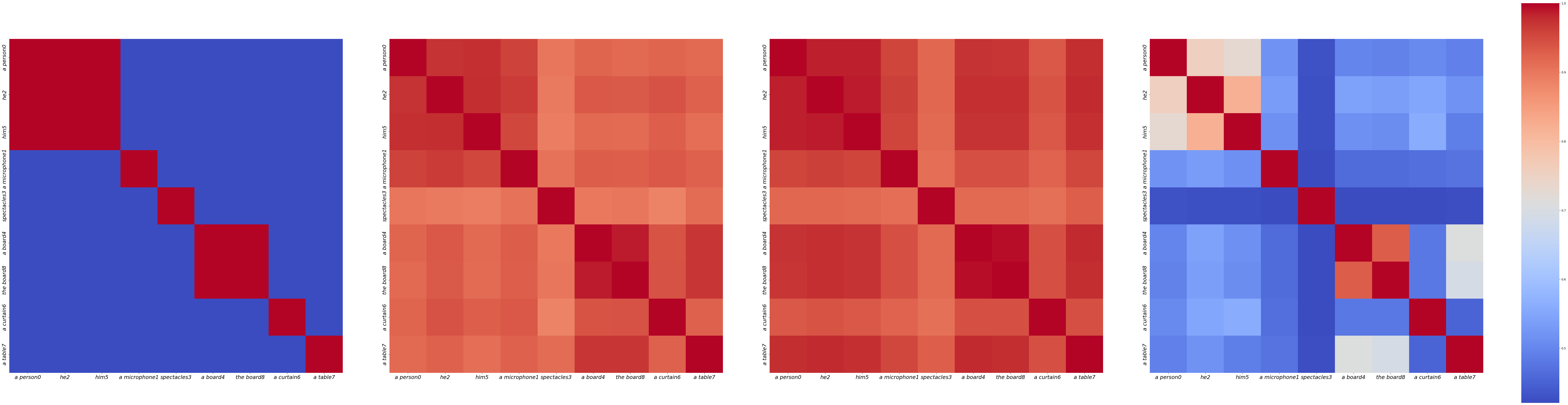}
\caption{Visualization of $f_{cor}(m,m^{\prime})$ for ground truth, visual-cue, category-cue, and PA-MCR model (left to right).}
\label{fig:ablation_visualization}
\end{figure}

\begin{table}[!t]
 \centering
    \caption{CR results (F1) on the CIN dataset with different fusion strategies. $\alpha$ denotes the weight of the visual cue.}
 \setlength{\tabcolsep}{8pt}
 \begin{tabular}{lcccc|c}
 \toprule
 \textbf{Strategy} & $\alpha$ & \textbf{MUC}& \textbf{B$^{3}$} & \textbf{CEAF$_{\phi4}$} & \textbf{CoNLL}\\
 \midrule
 \multirow{5}{*}{Strat. 1} & 0.2 & 37.41 & 90.09 & 86.42 & 71.31 \\
 & 0.4 & \underline{38.69} & 89.56 & 85.79 & 71.34 \\
 & 0.6 & 37.15 & 90.16 & \underline{86.51} & 71.28 \\
 & 0.8 & 36.64 & 89.91 & 86.17 & 70.91 \\
  \midrule
\multirow{5}{*}{Strat. 2} & 0.2 & 37.73 & 89.67 & 85.89 & 71.08 \\
 & 0.4 & 38.02 & 89.99 & 86.28 & 71.43 \\
 & 0.6 & 37.66 & 89.79 & 86.13 & 71.19 \\
 & 0.8 & 36.73 & 89.48 & 85.67 & 70.63 \\
  \midrule
\multirow{5}{*}{Strat. 3} & 0.2 & 37.81 & 89.61 & 85.85 & 71.09 \\
 & 0.4 & 38.10 & 90.01 & 86.35 & \underline{71.56} \\
 & 0.6 & 37.90 & 89.73 & 86.04 & 71.22 \\
 & 0.8 & 36.75 & \underline{90.16} & 86.46 & 71.12 \\
 \midrule
 \textbf{PA-MCR} & N/A & \textbf{39.28} & \textbf{90.31} & \textbf{86.48} & \textbf{72.02}\\
 \bottomrule
 \end{tabular}
     \label{tab:ablation_fusion}
 \end{table}

\subsection{Multi-cue Integration Study (RQ3)}\label{effectofcues}
\subsubsection{Effect of Different Cues}
We conducted an ablation study on the CIN dataset to assess PA-MCR with different cues (Table~\ref{tab:ablation}). ``Visual-cue" and ``Category-cue" denote the models formulated by  Eq.~(\ref{eq:matchingmentionregion}) but with $s_t(t^T, t^I)$ replaced by $s_t^v(t^T, t^I)$ and $s_t^c(t^T, t^I)$, respectively, while ``Multi-cue" denotes our full PA-MCR model. Results show that the multi-cue model consistently outperforms others. The category-cue variant surpasses the visual-cue variant, likely due to noise from redundant image regions detected visually. As illustrated in Fig.~\ref{fig:ablation_visualization}, visualization of $f_{cor}(m, m^{\prime})$ further shows that PA-MCR produces more structured and discriminative outputs, closely aligning with ground truth. These results demonstrate the superiority of our multi-cue integration strategy.
\subsubsection{Superiority to Other Cue Fusion Strategies}
To validate our ``express or suppress'' mechanism, we compared it with three alternative fusion strategies (Table~\ref{tab:ablation_fusion}): weighted summation of (1)~subject, predicate, and object embeddings; (2)~matching score matrices; and (3)~mention embeddings. Detailed descriptions of these fusion strategies can be found in Appendix~\ref{fusionStragies}. We replaced our integration with each strategy while keeping other components fixed. All three alternatives outperformed single-cue models with an appropriately selected weight, demonstrating multi-cue integration benefits. However, these methods rely on fixed cue-wise fusion weights, which cannot adapt to the varying reliability of different cues across instances. In contrast, evidence theory provides an uncertainty-aware fusion mechanism that dynamically adjusts the contribution of each cue on an instance-by-instance basis. As a result, even after exhaustive search over different weight configurations, the weighted-averaging-based alternatives still struggle with image-narration variability, whereas our method achieves superior performance. These results highlight the benefit of evidence-theoretic fusion.

\begin{figure}[!t]
    \centering
    \includegraphics[width=0.90\linewidth]{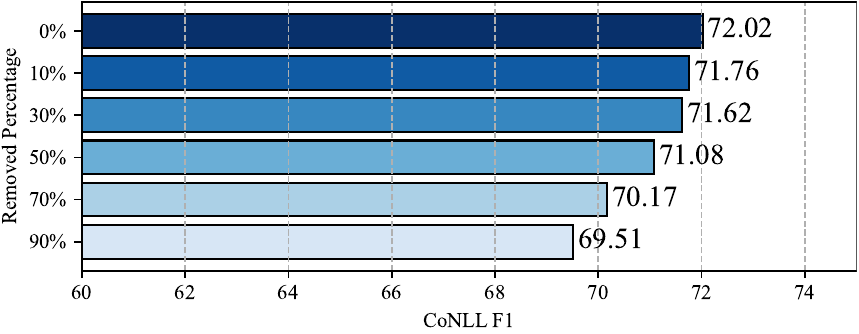}
    \caption{CoNLL F1 on the CIN dataset when randomly removing different percentages of generated textual triplets.}
    \label{fig:tripletremoved}
\end{figure}

\begin{figure}[!t]
 \centering
 \includegraphics[width=
 \linewidth]{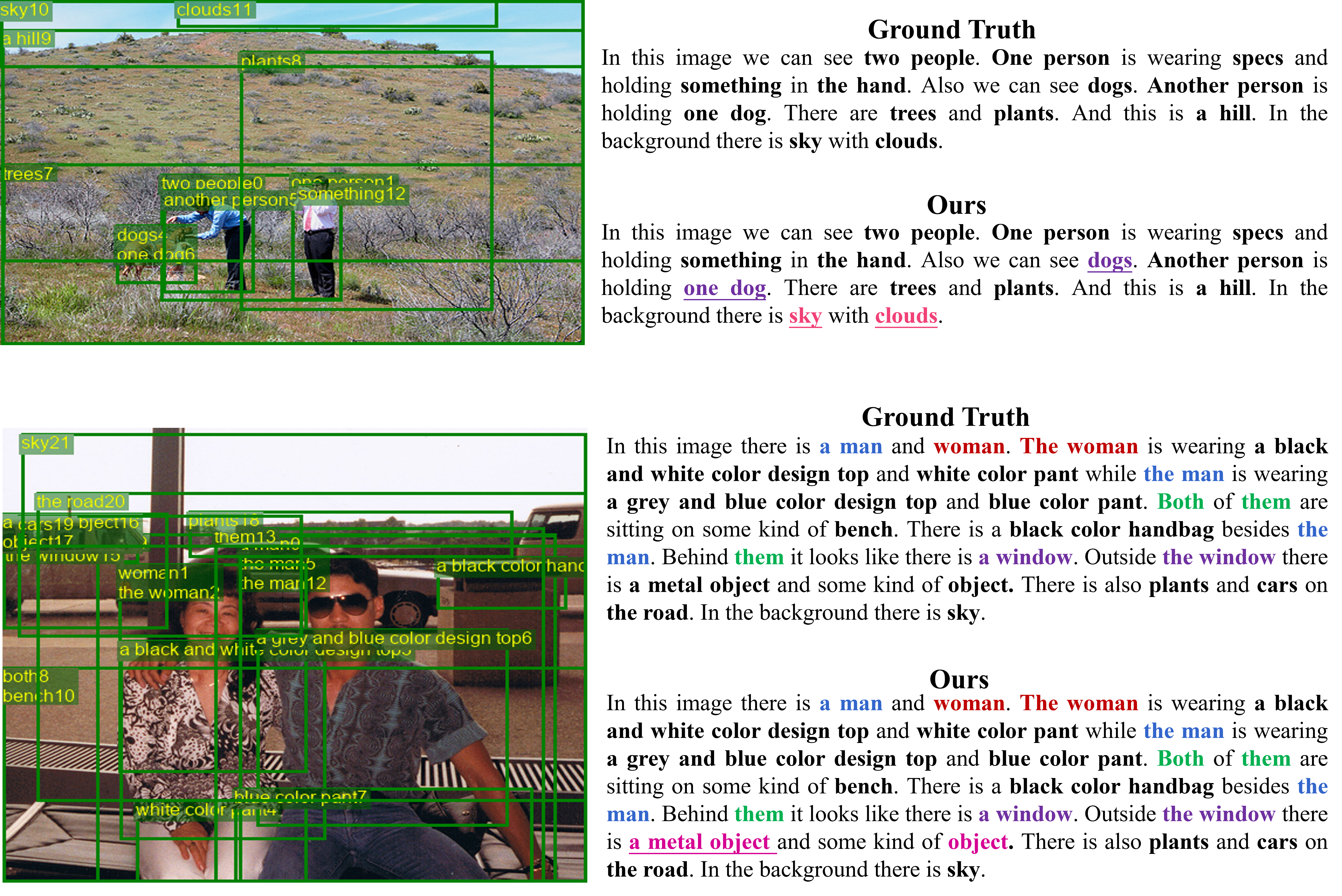}
 \caption{Qualitative Error Cases. Each row presents the grounding and coreference resolution (CR) results for a specific image narration. The left column shows the grounding results, and the right column displays the CR results. Mentions in the same color form a coreference chain, singletons (coreference chains of length one) are highlighted in bold for clarity, and miscoreferenced mentions are underlined.}
  \label{fig:qualitative_results}
\end{figure}

\subsection{Error Analysis (RQ4)} 

\subsubsection{Impact of Pre-Processing Errors}
A natural question is how the quality of pre-processing influences the final results. Directly measuring error propagation is challenging, as ground truth for upstream processing is unavailable. Although the masking experiments in Section~\ref{sec:Reliance} were not originally designed for this purpose, they provide an indirect illustration of how detection errors (i.e., undetected regions) propagate to the final results from the perspective of an object detector. Building on this, we evaluated the error propagation of the information extraction (IE) extractor by randomly removing varying percentages of generated textual triplets (CIN) associated with each mention\footnote{We ensure that each mention retains at least one associated textual triplet during the removal process.}. The results are presented in Fig.~\ref{fig:tripletremoved}. As expected, the performance of the IE extractor degrades progressively as more textual triplets are removed. However, the decline is relatively modest when only a small fraction of triplets is removed, indicating that PA-MCR can somewhat tolerate upstream IE errors. In Appendix~\ref{errors}, we provide the same error analysis for VR-CLIP, showing that the performance degradation of our method is more gradual and better controlled than that of strong baselines.

\subsubsection{Qualitative Analysis of Error Cases}
Although PA-MCR demonstrates strong performance on the coreference resolution task, it still exhibits several limitations. As illustrated in Fig.~\ref{fig:qualitative_results}, incorrectly predicted mentions are underlined for clarity. We observe that PA-MCR occasionally struggles to disambiguate mentions that are unrelated to others, such as \emph{dogs} in row 1. In this case, the model mistakenly links the mention to the orthographically similar \emph{one dog} in the same row, forming an incorrect coreference chain. The model also faces challenges when multiple mentions share the same contextual relationship. For example, in row 2, both \emph{a metal object} and \emph{object} are described as being ``outside the window", which leads to incorrect linking. Additionally, the model sometimes fails to differentiate between visually and semantically related entities, such as \emph{sky} and \emph{clouds}. 
To address these issues, performance could be improved by incorporating richer contextual information from both textual and visual modalities. One promising direction is to design a more effective contextual triplet extraction strategy that captures a broader and more nuanced context. Furthermore, explicitly modeling the interactions between text and image modalities during triplet extraction may help to resolve such ambiguities.

\subsubsection{A Close Look at Eq.~\eqref{eq:finalsimilarity}}
To better understand the model’s failure cases, we take a closer look at Eq.~\eqref{eq:finalsimilarity}, which directly determines the final mention-mention similarity. Although this weighted aggregation generally improves robustness, it may also mislead inference in certain extreme cases. In particular, consider two competing mention-region pairs, $(m,r_1)$ and $(m,r_2)$, that are associated with triplet pairs receiving similar matching scores. If the uncertainties of their corresponding triplet pairs are also close and relatively high, $q(t^T,t^I)$ approaches a nearly uniform scaling factor and loses its discriminative effect. As a result, both $s^\ast(m,r_1)$ and $s^\ast(m,r_2)$ are pulled toward similarly weighted averages of their respective triplet scores, causing the gap between them to shrink. This weakens region discrimination and can bias the downstream mention representation toward an incorrect coreference chain.

\section{Conclusion}
This paper presents PA-MCR, a novel multi-modal coreference resolver that operates effectively without requiring training on scarce benchmark datasets or relying on resource-intensive VLLMs. Unlike conventional approaches, PA-MCR leverages an existing alignment model and enhances it by incorporating relational information and multi-cue integration with evidence theory, yielding satisfactory performance. Extensive experiments on the CIN dataset demonstrate that PA-MCR achieves competitive performance compared to state-of-the-art dedicated methods and widely used VLLMs, highlighting its effectiveness despite not being trained on task-specific data. Furthermore, additional evaluations on masked CIN and VCR-MCR datasets confirm the method’s robustness and generalization capabilities. These results indicate that PA-MCR provides a promising benchmark-independent and accessible alternative for MCR.

\appendices
\section{Details of Other Cue Fusion Strategies}\label{fusionStragies}
In the multi-cue integration study, we investigate the effect of different fusion strategies. To make the comparison clearer, we provide a more detailed description of how each strategy is implemented. Let $(\mathbf{e}^{r_{k}},\mathbf{e}^{u_{k,l}},\mathbf{e}^{r_{l}})$ denote the embedding of an image triplet, $s_{t}(t^{T},t^{I})$ denotes the triplet-level matching score, and $s(m,r)$ denotes the similarity between mention $m$ and region $r$ under an arbitrary cue.
\subsection{Strat. 1: Fusion at the triplet-embedding level}
In this strategy, the visual and category cues are fused directly at the embedding level before triplet matching. Specifically, the fused triplet embedding is computed as
\begin{equation}
 \left\{
 \begin{array}{ll}
   \mathbf{e}_{k}^{*}=\alpha\mathbf{e}^{v_{k}}+(1-\alpha)\mathbf{e}^{c_{k}}, \\
   \mathbf{e}_{k,l}^{u_{*}}=\alpha\mathbf{e}^{v_{k,l}}+(1-\alpha)\mathbf{e}^{c_{k,l}}, \\
   \mathbf{e}_{l}^{*}=\alpha\mathbf{e}^{v_{l}}+(1-\alpha)\mathbf{e}^{c_{l}},
 \end{array}
 \right.
\end{equation}
where $(\mathbf{e}^{v_k}, \mathbf{e}^{v_{k,l}}, \mathbf{e}^{v_l})$, $(\mathbf{e}^{c_k}, \mathbf{e}^{c_{k,l}}, \mathbf{e}^{c_l})$, and $(\mathbf{e}_{k}^{*}, \mathbf{e}_{k,l}^{u_{*}}, \mathbf{e}_{l}^{*})$ denote the triplet embeddings under the visual cue, the category cue, and the fused representation, respectively, and $\alpha$ is the weight assigned to the visual cue.
\subsection{Strat. 2: Fusion at the triplet-matching-score level}
In this strategy, the visual and category cues are first used independently to compute triplet matching scores, and the resulting score matrices are then fused. Let $\mathbf{S}_{t}^{v} \in \mathbb{R}^{|\mathcal{T}^{T}|\times|\mathcal{T}^{I}|}$ denote the visual-cue triplet matching score matrix, where each element is $s_{t}^{v}(t^{T},t^{I})$. Similarly, let $\mathbf{S}_{t}^{c}$ denote the triplet matching score matrix under the category cue. The fused score matrix is defined as:
\begin{equation}
  \mathbf{S}_{t}^{*}=\alpha\mathbf{S}_{t}^{v}+(1-\alpha)\mathbf{S}_{t}^{c},
\end{equation}
where $\mathbf{S}_{t}^{*}$ is the fused triplet matching score matrix.
\subsection{Strat. 3: Fusion at the final mention-region similarity level}
In this strategy, mention-region similarities are independently computed under the two cues, and fusion is performed only at the final stage. Let $\mathbf{S}^{v} \in \mathbb{R}^{|\mathcal{M}|\times|\mathcal{R}|}$ denote the final mention-region similarity matrix under the visual cue, where each element is $s^{v}(m,r)$. Similarly, let $\mathbf{S}^{c}$ denote the corresponding matrix under the category cue. The fused similarity matrix is then given by:
\begin{equation}
  \mathbf{S}^{*}=\alpha\mathbf{S}^{v}+(1-\alpha)\mathbf{S}^{c},
\end{equation}
where $\mathbf{S}^{*}$ is the fused mention-region similarity matrix.

\begin{table}[!t]
\centering
\footnotesize
\caption{Comparison of the original formulation and the variant with explicit support for $\{\bar{A}\}$ on the CIN dataset}
\label{barA}
\begin{tabular}{lcccc}
 \toprule
\textbf{Method} & \textbf{MUC F1}& \textbf{B$^{3}$ F1} & \textbf{CEAF$_{\phi4}$ F1} & \textbf{CoNLL F1}\\
\midrule
PA-MCR($\{\bar{A}\}$) & 36.59 & 89.74 & 85.67 & 70.66 \\
\textbf{PA-MCR}          & \textbf{39.28} & \textbf{90.31} & \textbf{86.48} & \textbf{72.02} \\
\bottomrule
\end{tabular}
\end{table}

\begin{table}[!t]
\centering
\footnotesize
\caption{Comparison of the original formulation and the variant with $s_t(\cdot)=s_t^v(\cdot)$ on the CIN dataset}
\label{STV}
\begin{tabular}{lcccc}
 \toprule
\textbf{Method} & \textbf{MUC F1}& \textbf{B$^{3}$ F1} & \textbf{CEAF$_{\phi4}$ F1} & \textbf{CoNLL F1}\\
\midrule
$s_t(\cdot)=s_t^v(\cdot)$ & 37.55 & 90.27 & 86.33 & 71.39 \\
$s_t(\cdot)=s_t^c(\cdot)$ & \textbf{39.28} & \textbf{90.31} & \textbf{86.48} & \textbf{72.02} \\
\bottomrule
\end{tabular}
\end{table}

\begin{figure}[!t]
    \centering
    \includegraphics[width=0.9\linewidth]{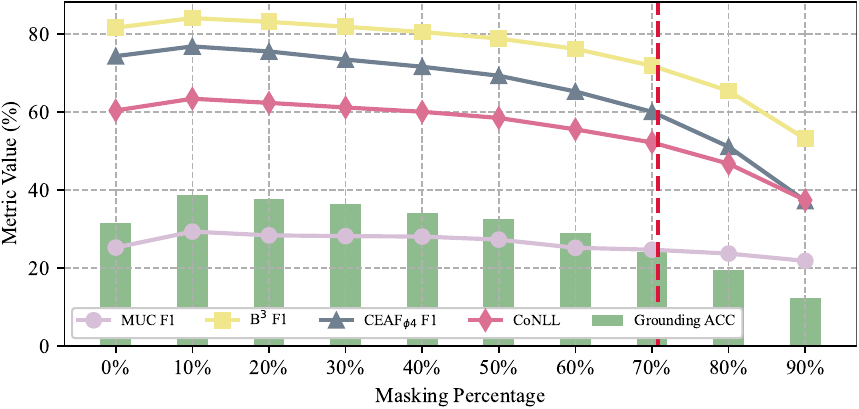}
    \caption{Grounding and CR results on the CIN dataset when randomly masking different percentages of image regions (VR-CLIP).}
    \label{fig:random_mask_VR}
\end{figure}

\begin{figure}[!t]
    \centering
    \includegraphics[width=0.9\linewidth]{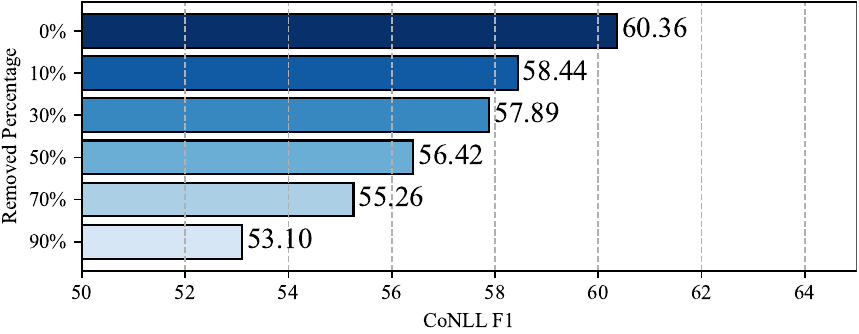}
    \caption{CoNLL F1 on the CIN dataset when randomly removing different percentages of generated textual triplets (VR-CLIP)}
    \label{fig:tripletremoved_VR}
\end{figure}

\section{Discussion of Key Formulations}\label{AAbS}
In this section, we provide a deeper discussion of several important design choices in our formulation.
\subsection{On Explicitly Modeling $\{\bar{A}\}$}
In our current formulation, we assign belief only to $\{A\}$ and the uncertainty set $U$, without explicitly modeling $\{\bar{A}\}$. The motivation is that low cosine similarity does not provide reliable negative evidence. Instead, it may simply reflect ambiguity or insufficient confidence. Therefore, we interpret low similarity as uncertainty rather than explicit evidence against matching.

A natural alternative is to use $1-S$ (where $S$ denotes matching similarity) as direct support for $\{\bar{A}\}$. To examine this possibility, we implement a variant, denoted as PA-MCR($\{\bar{A}\}$), in which $1-S$ is used to provide explicit support for $\{\bar{A}\}$. Table~\ref{barA} reports the comparison on the CIN dataset. Compared with the original formulation, this variant led to worse overall performance, which supports our design choice of treating low similarity as uncertainty rather than explicit negative evidence.
\subsection{On the Choice of $s_t(\cdot)$}
In Remark 2, we use the visual cue to support the category cue, and accordingly set $s_t(\cdot)=s_t^c(\cdot)$. The underlying motivation is that, although both cues originate from the visual side, the category cue is more semantically aligned with the textual modality. Compared with raw visual region features, category labels provide a more language-compatible representation, making them suitable as the primary cue in the evidential integration process. By contrast, raw visual region features contain abundant low-level visual details that are less directly aligned with textual semantics. The visual cue is therefore used as auxiliary evidence to support the category cue.

To further examine this design, we consider an alternative variant in which $s_t(\cdot)=s_t^v(\cdot)$. Table~\ref{STV} presents the comparison on the CIN dataset. The results show that directly using the visual cue leads to lower overall performance than using the category cue. This observation supports our original choice.

\section{Impact of Pre-Processing Errors of VR-CLIP}\label{errors}
To evaluate the impact of pre-processing errors on VR-CLIP, we conduct the same error analysis as for PA-MCR under degraded upstream model performance. The corresponding results are shown in Figs.~\ref{fig:random_mask_VR} and~\ref{fig:tripletremoved_VR}. Compared with PA-MCR, VR-CLIP exhibits weaker robustness to noisy or incomplete upstream evidence under the same relation-aware pre-training setting. This result further suggests that PA-MCR’s adaptation and fusion strategy help improve resilience to pre-processing errors.

\section{Prompt Template for VLLMs}\label{prompt}
This appendix provides the prompt template used for VLLMs in our experiments on MCR. The prompt is designed to guide the model to identify and cluster mentions that refer to the same underlying entity based on both visual evidence and textual narration. Specifically, we present below the task definition, detailed instructions, and the required output format.
\subsection{Task Definition}
\begin{tcolorbox}[colback=black!3,colframe=black!75,title=\textbf{Task}]
You are given an image, a narration describing the image, and a list of mentions. Your task is to perform coreference resolution over the mentions, i.e., group together only those mentions that refer to the same specific entity.
\end{tcolorbox}
\subsection{Step-by-step Instructions}
\begin{tcolorbox}[colback=black!3,colframe=black!75,title=\textbf{Instructions}]
\begin{enumerate}
    \item Use both the image and the narration to determine each mention’s referent.
    \item Group mentions together only when they refer to the same specific entity.
    \item Include every mention from the mentions list exactly once.
    \item If a mention does not corefer with any other mention, place it in its own singleton cluster.
    \item Copy each mention exactly as given in the mentions list.
    \item Do not add new mentions, omit mentions, or paraphrase mentions.
\end{enumerate}
\end{tcolorbox}

\subsection{Output Requirement}

\begin{tcolorbox}[colback=black!3,colframe=black!75,title=\textbf{Output}]
Return only a JSON object in the following format:

\begin{center}
\texttt{\{"clusters": \{"0": ["mention1", "mention2"], "1": ["mention3"]\}\}}
\end{center}

Each cluster should be represented as a list of mention strings.
\end{tcolorbox}

\bibliographystyle{IEEEtran}
\bibliography{refs}

\begin{IEEEbiography}[{\includegraphics[width=1in,height=1.25in,clip,keepaspectratio]{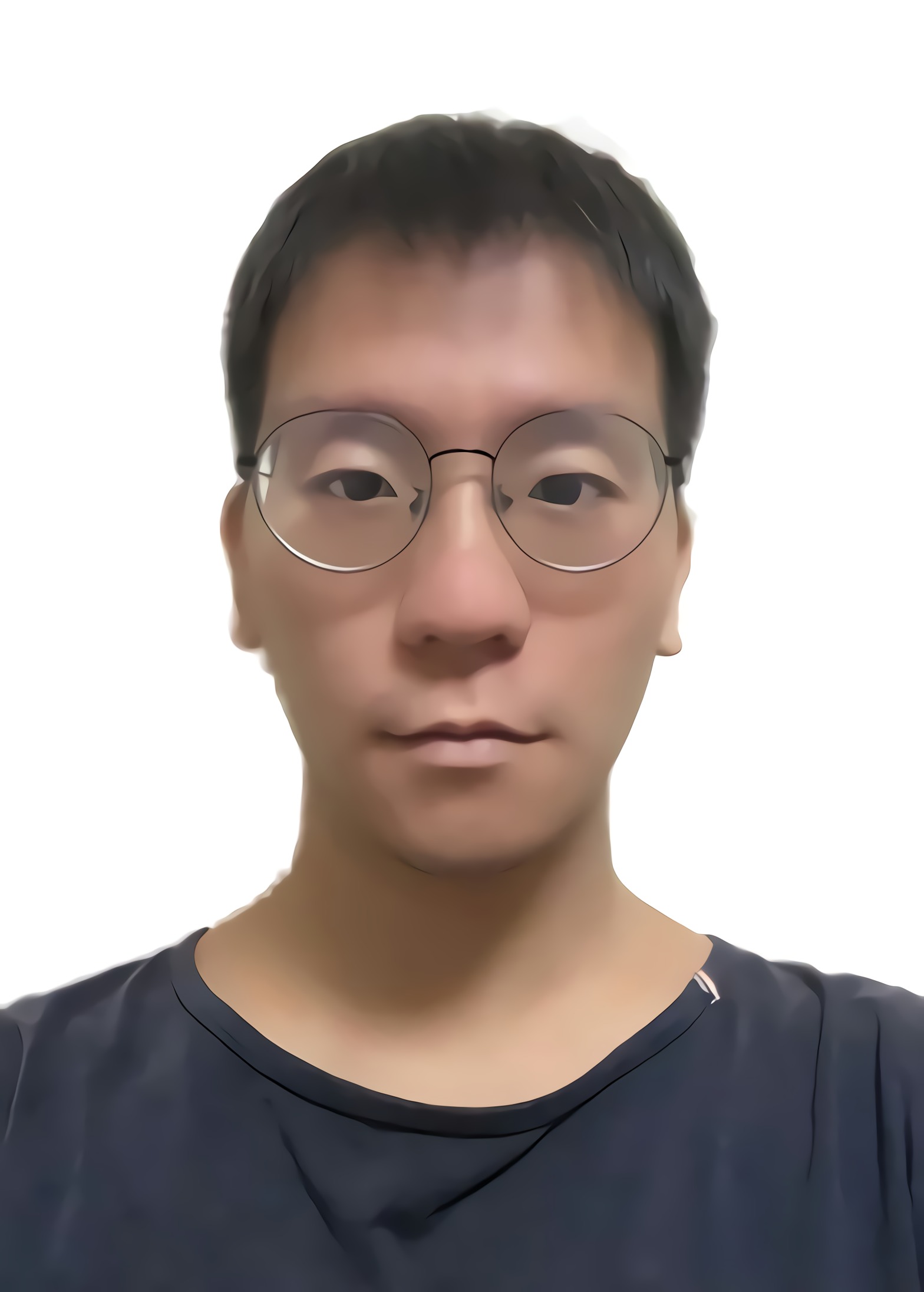}}]{Jinghan Wu}
received the B.S. degree in automation science and technology in Xi'an Jiaotong University, Shaanxi, China, in 2020. He visited the Centre for Frontier AI Research (CFAR), A*STAR, Singapore, from 2024 to 2025. Currently, he is pursuing his Ph.D. degree in the Institute of Artificial Intelligence and Robotics and College of Artificial Intelligence, Xi\rq an Jiaotong University. His current research interests include human-computer interaction, image processing, and machine learning.
\end{IEEEbiography}

\begin{IEEEbiography}[{\includegraphics[width=1in,height=1.25in,clip,keepaspectratio]{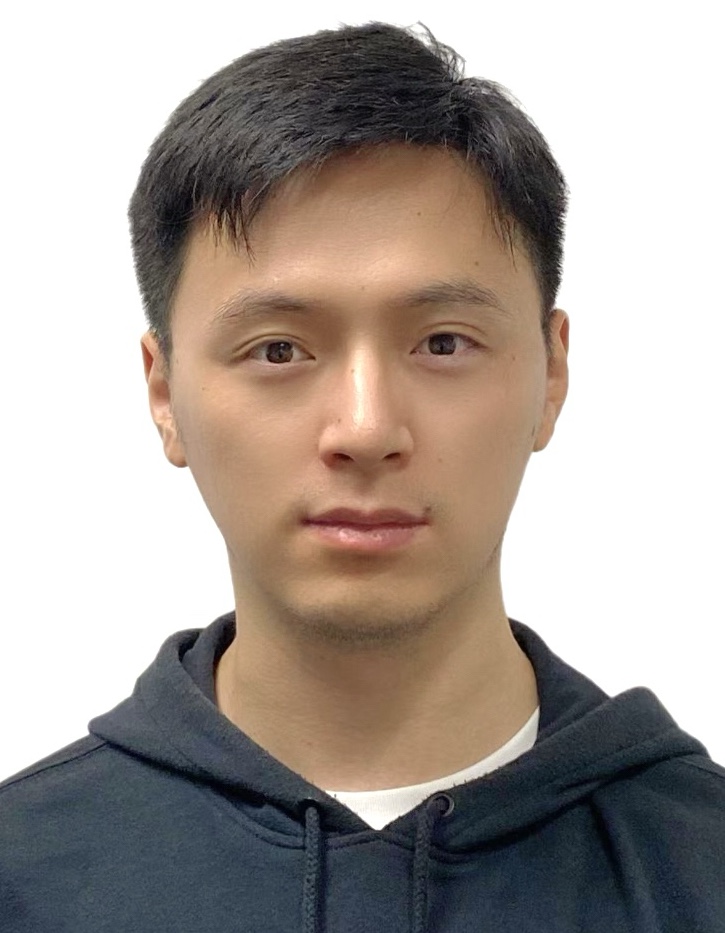}}]{Jing Li} is a Research Scientist at the Centre for Frontier AI Research (CFAR), A*STAR, Singapore. He received his Ph.D. in Information Technology from the University of Technology Sydney, Australia, in 2023, and his B.Eng. and M.Eng. degrees in Computer Science and Technology from Northwestern Polytechnical University, Xi’an, China, in 2015 and 2018, respectively. His current research focuses on trustworthy machine learning.
\end{IEEEbiography}

\begin{IEEEbiography}[{\includegraphics[width=1in,height=1.25in,clip,keepaspectratio]{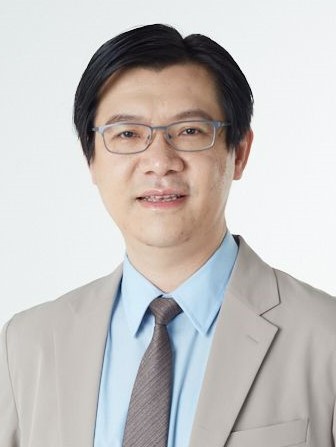}}]{Ivor W. Tsang} is the Director of A*STAR Centre for Frontier AI Research. He is an Adjunct Professor at School of Computer Science and Engineering (SCSE), Nanyang Technological University, Singapore. His research interests include transfer learning, deep generative models, learning with weakly supervision, Big Data analytics for data with extremely high dimensions in features, samples and labels. In 2013, he was the recipient of the ARC Future Fellowship for his outstanding research on Big Data analytics and large-scale machine learning. In 2019, his JMLR paper Towards ultrahigh dimensional feature selection for Big Data was the recipient of the International Consortium of Chinese Mathematicians Best Paper Award. In 2020, he was recognized as the AI 2000 AAAI/IJCAI Most Influential Scholar in Australia for his outstanding contributions to the field between 2009 and 2019. His research on transfer learning granted him the Best Student Paper Award at CVPR 2010 and the 2014 IEEE TMM Prize Paper Award. He is an IEEE Fellow, and serves on the Editorial Board for the JMLR, MLJ, JAIR, IEEE TPAMI, IEEE TAI, IEEE TBD, and IEEE TETCI. He serves/served as an AC or Senior AC for NeurIPS, ICML, AAAI and IJCAI, and the steering committee of ACML.
\end{IEEEbiography}

 \begin{IEEEbiography}[{\includegraphics[width=1in,height=1.25in,clip,keepaspectratio]{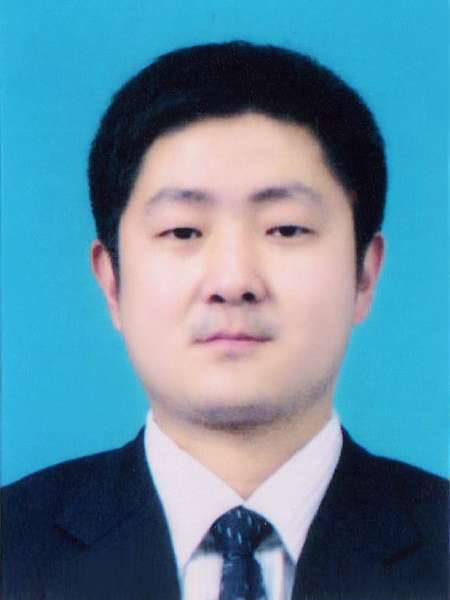}}]{Xuetao Zhang}
received the B.S. degree in information engineering, M.S. degree and Ph.D. degree in pattern recognition and intelligence system from Xi\rq an Jiaotong University, China, in 2003, 2006 and 2012 respectively. He visited the Department of Brain and Cognitive Sciences, Massachusetts Institute of Technology from 2009 to 2010. He is currently a professor at the Institute of Artificial Intelligence and Robotics in Xi\rq an Jiaotong University. His research interests include computer vision, human vision, and machine learning.
\end{IEEEbiography}

\end{document}